\documentclass[sigconf]{acmart}

\AtBeginDocument{%
  \providecommand\BibTeX{{%
    \normalfont B\kern-0.5em{\scshape i\kern-0.25em b}\kern-0.8em\TeX}}}

\usepackage{xspace}
\usepackage{multirow}

\newcommand{\distmult}{\textsc{DistMult}\xspace}
\newcommand{\rgcn}{\textsc{RGCN}\xspace}
\newcommand{\szat}{\textsc{Comp}\xspace}

\newcommand{\relatt}{\textsc{RelAtt}\xspace}
\newcommand{\wordnet}{\textsc{WordNet}\xspace}
\newcommand{\fbk}{\textsc{FB15k-237}\xspace}
\newcommand{\wn}{\textsc{WN18}\xspace}

\setcopyright{acmcopyright}
\copyrightyear{2020}
\acmYear{2020}
\acmDOI{10.1145/1122445.1122456}

\acmConference[San Diego '20]{San Diego '20: 2nd International Workshop on Deep Learning on Graphs: Methods and Applications}{August 23--27, 2020}{San Diego, CA}
\acmBooktitle{San Diego '20: 2nd International Workshop on Deep Learning on Graphs: Methods and Applications,
  August 23--27, 2020, San Diego, CA}

\begin{document}

\title{Knowledge Graph Embedding using Graph Convolutional Networks with Relation-Aware Attention}

\author{Nasrullah Sheikh}
\authornote{Both authors contributed equally}
\affiliation{%
  \institution{IBM Research - Almaden}
  \city{San Jose}
  \state{CA}
}

\email{nasrullah.sheikh@ibm.com}
 \author{Xiao Qin}
  \authornotemark[1]
\affiliation{%
  \institution{IBM Research - Almaden}
  \city{San Jose}
  \state{CA}
}
\email{xiao.qin@ibm.com}

\author{Berthold Reinwald}
\affiliation{%
  \institution{IBM Research - Almaden}
  \city{San Jose}
  \state{CA}}
\email{reinwald@us.ibm.com}

\author{Christoph Miksovic}
\affiliation{%
  \institution{IBM Research - Zurich}
  \state{Switzerland}}
  
\email{cmi@zurich.ibm.com}
 \author{Thomas Gschwind}
\affiliation{%
  \institution{IBM Research - Zurich}
  \state{Switzerland}}
\email{thg@zurich.ibm.com}

\author{Paolo Scotton}
\affiliation{%
  \institution{IBM Research - Zurich}
  \state{Switzerland}}
\email{psc@zurich.ibm.com}

\renewcommand{\shortauthors}{Nasrullah, et al.}

\begin{abstract}
Knowledge graph embedding methods learn  embeddings of entities and relations in a low dimensional space which can be used for various downstream machine learning tasks such as link prediction and entity matching. Various graph convolutional network methods have been proposed which use different types of information to learn the features of entities and relations. However, these methods assign the same weight (importance) to the neighbors when aggregating the information, ignoring the role of different relations with the neighboring entities. To this end, we propose a relation-aware graph attention model that leverages relation information to compute different weights to the neighboring nodes for learning an embeddings of entities and relations. We evaluate our proposed approach on link prediction and entity matching tasks. Our experimental results on link prediction on three datasets (one proprietary and two public) and results on unsupervised entity matching on one proprietary  dataset demonstrate the effectiveness of the relation-aware attention. 


\end{abstract}



\maketitle

\section{Introduction}
\begin{figure}
     \centering
     \includegraphics[width=\linewidth]{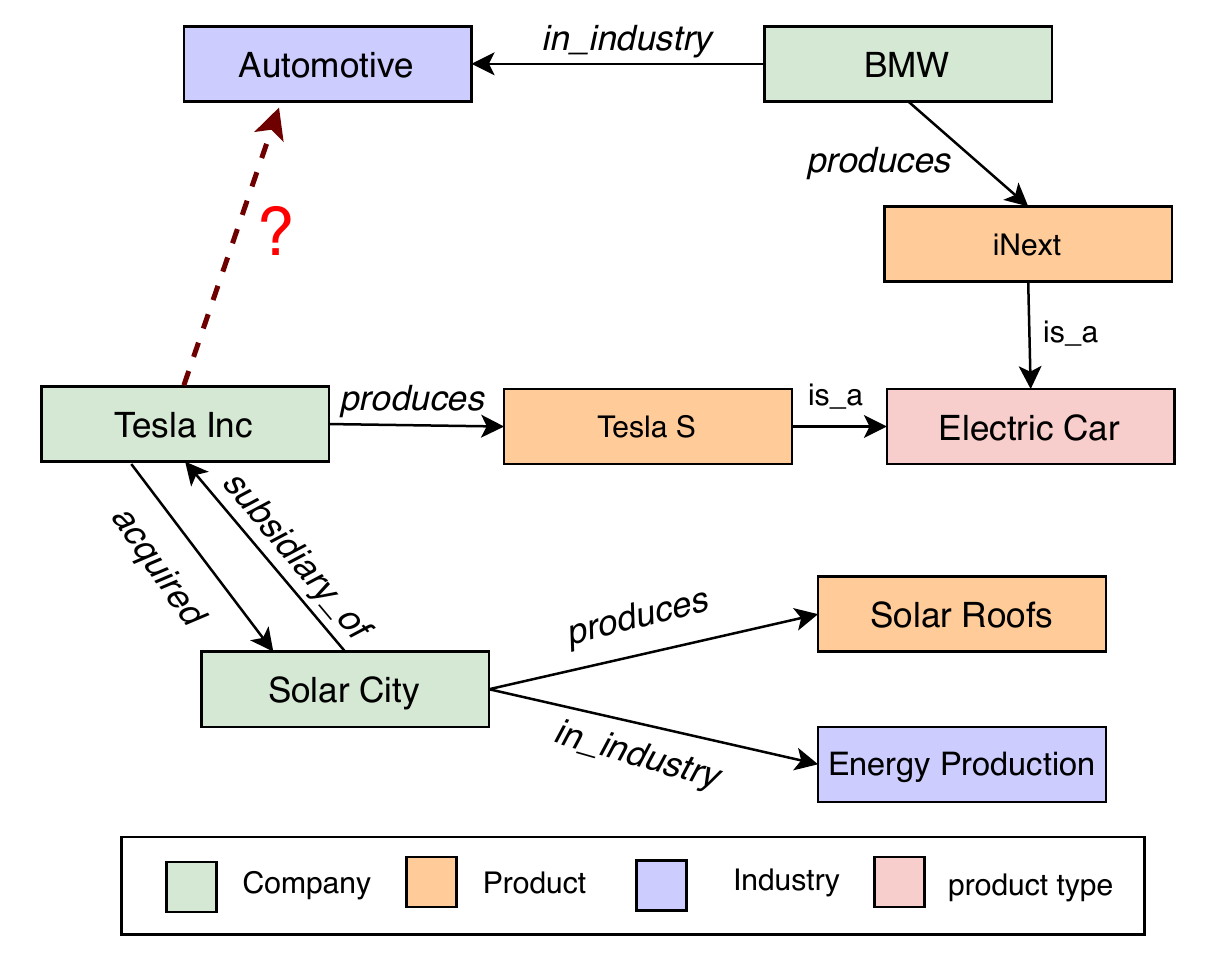}
     \caption{An example of an enterprise Knowledge Graph having 3 entity types and 5 link types. The red dotted arrow is the missing link information. }
     \label{fig:kg}
 \end{figure}

Knowledge Graphs (KG)  represent facts in the form of entities and relations between them. A fact is represented by a triplet $(h,r,t)$ where $h$, $t$ represent the head and tail entities respectively, and $r$ represents the relation between $h$ and $t$. Furthermore,  entities and relations may have some additional information such as attributes associated with them. Data from  different domains such as enterprises, gene ontology, etc.  can be modeled as KGs which is useful in different applications. KGs are critical to enterprises as they enable an organization to view, analyze, derive inferences, and build up knowledge for competitive advantage; for example, discovering new links between entities is useful in many scenarios such as discovering new side effects of a drug, or establishing new corporate relationships. One of the biggest challenges is to extract the data from various  structured and unstructured sources and build it in a KG such that it can be used effectively in various tasks such as search and answering, entity matching, and link prediction. 
An example of an enterprise knowledge graph is shown in Figure~\ref{fig:kg}. The graph shows companies, subsidiaries, products, industry types, and product types represented as entities;   \textit{produces}, \textit{in\_industry},  \textit{subsidiary\_of}, \textit{acquired}, and \textit{is\_a} represent the relationships between the entities. 
 
Often these KGs are sparse and have missing information. For example, in Figure~\ref{fig:kg}, the relation between \textit{Tesla Inc} and \textit{Automotive} is missing ($\textit{<Tesla Inc, ?, Automotive>}$). In general, the missing information in KGs can be of the form $(h,r,?)$, $(?,r,t)$, and  $(h,?,t)$. Towards this end, various methods have been proposed such as ~\cite{distmult, conve, DBLP:conf/esws/SchlichtkrullKB18, MLP, complex} which learn embeddings of entities and relations, and use a scoring function to determine if a triplet $(h,r,t)$ is valid or not. 
Models such as DistMult~\cite{distmult} process each triplet independent of other triplets, and hence do not exploit the neighborhood information in learning. Graph convolution based methods overcome this problem by aggregating the features from the neighboring entities and applying a transformation function to compute the new features. But these graph-based methods give equal weights to each of the neighboring entities, ignoring that the neighbors have different significance in computing new features~\cite{velickovic2018graph}. This attention mechanism considers edges having the same type, thus, it  cannot be directly extended to knowledge graphs which have multiple relation types between entities.

In a knowledge graph, relation types between entities determine the semantics of an edge. This semantic information is crucial in various downstream tasks such as link prediction and entity matching. Therefore, the relationship types cannot be ignored in computing the importance of neighbors. Towards this end, we introduce \relatt -- a relation aware masked attention mechanism in knowledge graphs which includes the features of relation for computing the attention. This attention is applied to the messages from neighbors during the propagation phase of graph neural networks to learn the embedding of entities and relations. The learning is optimized through a scoring function which scores a valid triplet higher than an invalid triplet based on the representation of entities and relations. Moreover, our proposed model is inductive, i.e., the learned model can be used to infer embeddings of unseen nodes. 
We evaluate our proposed approach on link prediction using two public datasets and one proprietary dataset against the state-of-the-art baselines. We also evaluate the embeddings on unsupervised entity linking task on the proprietary dataset. 





The rest of the paper is organized as follows: Section~\ref{sec:relwork} presents the related work, and Section~\ref{sec:bacg} provides a brief description of  Graph Neural Networks. In Section~\ref{sec:model}, we describe our proposed model, Section~\ref{sec:training} describes the procedure to train the model. In Section~\ref{sec:expeval}, we describe the evaluation on link prediction task (Subsection~\ref{subs:linkpre}) and entity matching task (Subsection ~\ref{subs:entitym}) which includes  details of datasets, experimental settings and results respectively.  Finally, we conclude in Section~\ref{sec:conc}.

\section{Related Work}
\label{sec:relwork}
Knowledge graph embedding methods have drawn a lot of attention due to recent advancements in representation learning. KG embedding methods learn a representation of entities and relations, and these representations are used for various downstream machine learning tasks such as KG Completion~\cite{transh} and entity classification~\cite{DBLP:conf/esws/SchlichtkrullKB18}.
Knowledge graph embedding methods for link prediction can be classified into two groups: \textit{translational and semantic models} and \textit{neural network models}. Moreover, these methods either use only the observed facts or exploit additional information to learn better embeddings such as entity types~\cite{10.5555/3060832.3061036} and  logical rules~\cite{10.5555/2832415.2832507}. 
\paragraph{Translational and Semantic Models} use a different parameterization of entities or relations, and scoring functions to determine the plausibility of a fact in a knowledge graph. Given a triplet $(h,r,t)$, translation-based methods~\cite{bordes2013translating,transr,transh}  use relation as a translational vector and apply a translation by relation on $h$. For a fact to hold, the embedding of translated entity $h$ should be close to $t$ i.e, the distance between these two should be minimum. TransE~\cite{bordes2013translating}  represents entities and relations in same embedding space and uses a margin-based distance scoring function. TransE cannot handle one-to-many and many-to-many relations. To overcome this, various other translation-based methods have been proposed such as ~\cite{transh, transr}. TransR~\cite{transr} represents each relation in a different relational space and both entities $(h,t)$ are first translated into the relation space and then distance function is applied to check the validity of a triplet. On the other hand, semantic-based models~\cite{distmult,rescal,complex} classify triplets based on a similarity function. These models fail to capture the complex relationships and are also limited to learn expressive features due to  their shallow structure. 
\paragraph{Neural Models} - use neural network models~\cite{DBLP:conf/esws/SchlichtkrullKB18,conve, convkb, MLP, kbgan}  to learn better embeddings of entities and relations which are then subsequently used in downstream tasks. Dong et al.~\cite{MLP} proposed a multi-layer perceptron based approach where embeddings of $h,r,t$ are concatenated at the input layer and non-linear transformation is applied to classify the triplets.  Convolutional Neural Network-based approaches have been proposed such as ConvKB~\cite{convkb} and ConvE~\cite{conve}. The entities and relations in ConvKB and ConvE are represented in 1D and 2D respectively, and both employ a convolutional operation to learn  embeddings of entities and relations, which are used by a scoring function for classification of triplets.  Schlichtkrull et al.~\cite{DBLP:conf/esws/SchlichtkrullKB18} proposed \rgcn~- a relational graph convolutional neural network model which uses a message-passing approach to aggregate the features from neighboring entities to learn the embedding of entities and relations to use in entity classification and link prediction tasks. These models give equal weights to the neighbors of a node which limits the learning of quality embeddings because the neighborhood nodes have different relationships and significance.

\section{Background}
\label{sec:bacg}
For the sake of completeness, in this section we provide a brief description of message passing based Graph Neural Networks (GNNs). For each node, a message passing GNN~\cite{gin} iteratively aggregates representation from its neighbors. Each iteration defines a layer of GNN, and $l$ iterations (layers) encode the structural information of the graph within its $l$-hop neighborhood. The $l$-th layer of a GNN is described as:

\begin{equation}
\label{eq:gnn}
a^{(l+1)}_{v_i} = \textsl{Agg}\big(\big\{h^{(l)}_u | u\in \mathcal{N}(v_i)\big\}\big),
\end{equation}


\noindent where $h^{(l+1)}_{v_i}$ is the vector representation of the node $v_i$ at the $l$-th iteration. The $\mathcal{N}(.)$ function returns neighbors of a node, and $\textsl{Agg}$ is the aggregation function which is defined as per the modelling approaches of different methods~\cite{DBLP:conf/esws/SchlichtkrullKB18,hamilton2017inductive}.
\section{Methodology}
\label{sec:model}
In this section, we describe \relatt (\underline{Rel}ation-aware \underline{Att}ention), model  that  exploits the relation between two entities to learn the importance of neighboring nodes by using the idea of ~\cite{velickovic2018graph}, and then recursively propagates node features in the graph. Our model has two components: \textit{embedding layer}, and  
\textit{knowledge graph convolutional layer with relation-aware attention}

\subsection{Embedding Layer}
The additional information such as attributes associated with entities in the KG contain semantic information and thus, can be leveraged in learning. 
For each entity, we concatenate the various textual attributes and use a pre-trained BERT~\cite{bert} model to obtain their  embeddings. These embeddings form an initial feature vector of entities to be used in the training. 
In case of datasets which do not have attributes, the embedding layer is initialized randomly.

\subsection{Knowledge Graph Convolutional Layer  with Relation-Aware Attention}
This layer  is  defined  as a single neural network layer which performs \textit{relation-aware attention}, \textit{feature propagation and aggregation}. 
The input to this layer is set of $N$ node features from embedding layer, $\mathbf{h} = \{\mathit{h}_1, \mathit{h}_2, \cdots , \mathit{h}_N\}$ where $\mathit{h}_i \in \mathbb{R}^d$ represents the $d$-dimensional features of $i^{th}$ node; a set of relation types $R = \{r_1, r_2, \cdots, r_k\}$; and  a set of relation features $\mathbf{m} = \{m_1, m_2, \cdots m_k \}$, where $m_r \in \mathbb{R}^d$ is the feature vector of $r^{th}$-relation type of dimension $d$.

 \begin{figure}[t]
     \centering
     \includegraphics[scale=0.65]{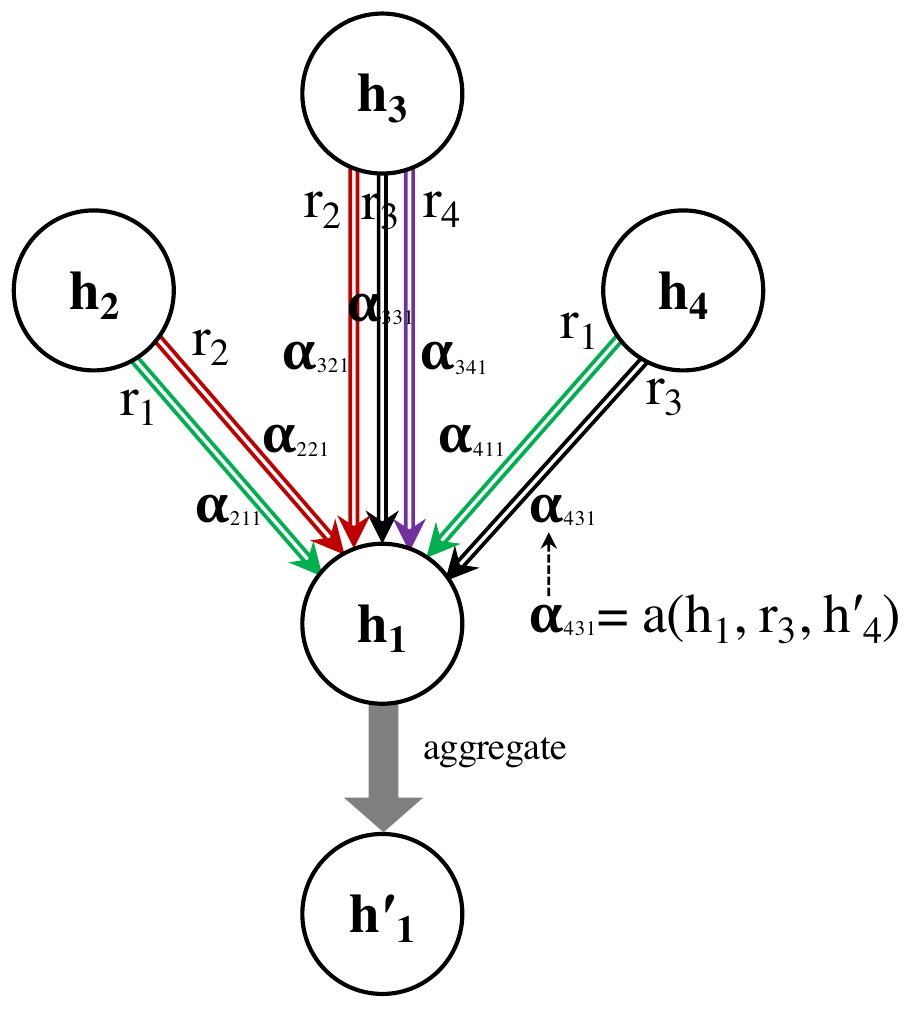}
     \caption{The attention mechanism of \relatt.}
     \label{fig:att}
 \end{figure}

\subsubsection{Relation-Aware Attention}
In a  Knowledge Graph, nodes have different types of relationships, thus the importance of neighbors is not only dependent  on their features but also on the features of relationships. 
To this end, we follow \cite{velickovic2018graph} and apply a shared linear transformation on triplets $(h,r,t)$, 
parameterized by a weight matrix $\mathbf{W}$. Then, we  perform \textit{self-attention} with respect to a shared relation between entities to compute attention coefficients: $a: \mathbb{R}^d \times \mathbb{R}^d \times \mathbb{R}^d \rightarrow \mathbb{R}$. The attention mechanism is shown in Figure~\ref{fig:att}. The attention coefficient  of a triplet $(h,r,t)$ is computed as: 
\begin{equation}
e_{(h,r,t)} = a(\mathbf{W}h_h, \mathbf{W}m_r, \mathbf{W}h_t)
\end{equation}

The attention mechanism $a$ is a trainable function parameterized by a weight vector $\mathbf{a} \in \mathbb{R}^{3d}$, which is given as:
\begin{equation}
\label{eq:attn:coeff}
    e_{(h,r,t)} = \mathbf{a}^T[\mathbf{W}h_h || \mathbf{W}m_r || \mathbf{W}h_t]
\end{equation}
where $.^T$ and $||$ are transpose and concatenation operations respectively. Moreover, the attention is masked i.e, the attention is computed for directly connected neighbors only given by $\mathcal{N}_h$. 
We apply softmax to make attention coefficients comparable across the neighborhood as given in Eq.~\ref{eq:attn:sftmax}
\begin{equation}
\label{eq:attn:sftmax}
    \alpha_{(h,r,t)} = softmax (e_{(h,r,t)}) = \frac{exp(e_{(h,r,t)})} {\sum_{(r^\prime,t^\prime) \in \mathcal{N}_{h}} exp(e_{(h,r^\prime,t^\prime)})} 
\end{equation}

\subsubsection{Feature Propagation and Aggregation}
Knowledge graph convolutional network architectures such as \rgcn~\cite{DBLP:conf/esws/SchlichtkrullKB18} consider the heterogeneity of the edges and  use a message passing framework to compute a new representation of a head node by  applying some relation-specific transformation on representation of neighbors before aggregating at the head node.
Following Equation~\ref{eq:gnn}, a generalized framework for knowledge graphs can be expressed as:
\begin{equation}
\label{eq:mpass}
    h_h^{\prime} = \sigma \left( \textsl{Agg}_{(r,t) \in \mathcal{N}_h} f(h_h, r, h_t)\right)
\end{equation}
where $f$ is a relation-specific transformation on representation of immediate neighborhood nodes given by $\mathcal{N}_h$, $\textsl{Agg}$ is an aggregator function such as \textit{SUM}, \textit{MEAN}  that combines these transformed messages from neighbors before passing it to an activation function $(\sigma)$, and $h_{h}^\prime$ is the new hidden features of entity $h$.

Combining Equation~\ref{eq:mpass} and Equation~\ref{eq:attn:sftmax} describes a single neural layer for knowledge graph convolution with relation-aware attention.
\begin{equation}
\label{eq:mpass:attn}
    \mathit{h}_h^{\prime} = \sigma \left( \textsl{Agg}_{(r,t) \in \mathcal{N}_h} \alpha_{(h,r,t)} f(\mathit{h}_h, \mathit{r}, \mathit{h_t})\right)
\end{equation}

The Equation~\ref{eq:mpass:attn} is agnostic to the underlying knowledge graph convolution message passing paradigm. However, in this work we use \rgcn~\cite{DBLP:conf/esws/SchlichtkrullKB18} as the underlying convolutional message passing method. Following \rgcn which uses \textit{SUM} as aggregation function, we can extend Equation~\ref{eq:mpass:attn} to $L$-layers as:
\begin{equation}
\label{eq:rgcn:attprop}
    \mathit{h}_h^{(l+1)} = \sigma \left (\sum_{r \in R} \sum_{t \in \mathcal{N}_h^r}
    \alpha_{(h,r,t)}
    \frac{1}{\mathcal{N}_h^r}\mathbf{W}_r^{(l)}h_t^{(l)} + \mathbf{W}_0^{(l)}h_h^{(l)} \right)
\end{equation}
where $\mathbf{W}_r^{(l)}$ is the weight matrix corresponding to relation $r$ in $l^{th}$-layer, and $\mathcal{N}_h^r$ gives the set of neighbors which share relation $r$ with entity $h$.


\section{Training}
\label{sec:training}
The objective of the knowledge graph based embedding methods is to learn  embeddings of entities and relations which are fed into a scalar output producing scoring function$(g)$ which scores true triplets much higher than false triplets. Various methods such as \distmult~\cite{distmult}, NTN~\cite{NTN} have proposed different scoring functions.  Our proposed method given above is limited to use only scoring functions which describe relation in $\mathbb{R}^d$ space. Therefore,  we use \distmult~\cite{distmult} scoring function given as:
\begin{equation}
    \label{eq:scoringfun}
    g(h,r,t) = h_h^T\,\mathbf{M}_r\,h_t
\end{equation}
We train the model using a negative sampling~\cite{distmult,DBLP:conf/esws/SchlichtkrullKB18} approach. For each positive triplet $\tau \in T^+$, we generate a set of negative samples by either corrupting $h$ or $t$ which produces a set of negative triplets $T^-$. Given the set of positive and negative triplets $T=T^+ \bigcup T^-$,  we optimize the model on cross entropy loss so as to learn entity and relation embeddings.

\begin{equation}
    \label{eq:loss}
    \mathcal{L} = \frac{1}{|T|} \sum_{\tau \in T} y~\textsl{log}~l\big(g(
    \tau)\big) + (1-y)\textsl{log}\big(1- l(g(\tau))\big) 
\end{equation}
where $\tau$ is training example $(h,r,t)$; $l$ is logistic \textit{sigmoid} function; $y$ is 1 or 0 for  positive or negative triplet respectively.

\section{Experimental Setup and Evaluation}
\label{sec:expeval}
In this section, we provide the details of datasets, baseline methods for comparison, and training settings and evaluation protocol for link prediction and entity matching tasks.

\subsection{Supervised Task: Link Prediction}
\label{subs:linkpre}
First, we evaluate \relatt with the link prediction task in a supervised learning setting. That is, the \relatt is trained using a large portion of the original knowledge graph, and the goal is to predict the missing $h$ or $t$ in the omitted triplets. Link prediction is a common task for evaluating a knowledge graph embedding method and is typically measured by Mean Reciprocal Rank (MRR) and Hits@K. 
For each test triplet, we obtain a set of all possible triplets and score them through the model. The triplets are ranked on the score, and the position of true test triplet in a sorted list is its rank($c$), and $1/c$ is reciprocal rank. The mean of reciprocal ranks of all true test triplets is called Mean Reciprocal Rank. Hits@K gives the number of times the test triplets occur in top $k$ rankings in the ranked list. The higher values of MRR and Hits@K indicate the better performance of the model.  
\subsubsection{Datasets}
We evaluate our proposed approach on three datasets - two widely used public datasets (\fbk and \wn)  and one proprietary dataset (\szat). \fbk~\cite{toutanova-chen-2015} is obtained from FB15k by removing inverse relations; which is a subset of relational database FreeBase, containing general facts.  \wn~\cite{bengioKB} is a subset of \wordnet and contains lexical relations between words. It mostly contains \textit{hyponym} and \textit{hypernym} relations. \szat is a proprietary dataset and is  extracted from the relational database of companies which include different branches, subsidiaries, headquarters, and products manufactured/produced by companies located in two countries. The companies, products and cities are the entities, and there are 6 different relationships between these entities\footnote{\textit{manufactures, category\_of, parent\_category, owns, in\_city, in\_country}}. The train, test and validation splits are from \distmult~\cite{distmult} and in case of \szat dataset, we used 80\% triplets for training, and  10\% of triplets each for validation and test. The statistics of the datasets are given in Table~\ref{tab:data}.


\begin{table}[t]
\caption{Dataset statistics.}
\label{tab:data}
\begin{tabular}{cccc}
\toprule
    Dataset      & \wn       & \fbk    & \szat \\
    \midrule
\# Entities       &  40943       &    14,541     &   11,585   \\
\# Relations     &   18         &   237         &    6  \\
\# Features      &   -          & -             & 768 \\
\# Train Triplets &   141,442    &   272,115     &     60,177 \\
\# Valid Edges   &  5000        &    17,535       &   7,522   \\
\# Test Edges     &  5000        &     20,466      &     7,522 \\
\bottomrule
\end{tabular}
\end{table}


\subsubsection{Baselines}
We compared our approach with two methods: \distmult~\cite{distmult} and \rgcn~\cite{DBLP:conf/esws/SchlichtkrullKB18}. \distmult is a factorization based method which represents entities in a $d$ dimensional vector space and each relation is represented by a  diagonal matrix . \rgcn is a graph convolutional based approach to learn embeddings of entities and relationships. For experimental evaluation we used DGL\cite{wang2019dgl} of \rgcn, and HNE\cite{yang2020heterogeneous} implementation of \distmult. For the dataset with attributes, we used a variant \rgcn that includes an embedding layer for attributes. 
Since, \distmult does not use attribute information, therefore we omitted the attributes for learning on the \szat dataset . 

\subsubsection{Experimental Setup and Results}
\label{subs:expsetup}
We selected hyperparameters of our model and baselines on their respective validation sets by grid search and early stopping on the filtered Mean Reciprocal Rank (MRR) metric using full-batch optimization. For fairness of comparison, we used the same hyperparameter search space for our model and the baselines, and trained  models for a minimum of 6000 epochs. The  hyperparameter space is given as - learning rate \{0.01, 0.001\}, number of hidden layers \{1,2\}, hidden layer dropout \{0.0,0.1,0.2, 0.3\}, attention dropout \{0.0,0.1,0.3,0.6\}, embedding size \{100, 200,400\}, number of bases \{2,3,5,10,50,100\}, and  negative samples \{10\}. We used \textit{basis decomposition}~\cite{DBLP:conf/esws/SchlichtkrullKB18} for regularization and optimized the loss function using Adam optimizer.

After training the models, we follow the evaluation protocol of Yang et.al ~\cite{distmult} for link prediction and report results on two widely used and standard metrics: \textit{Mean Reciprocal Rate} and \textit{Hits@k} (1,3,10). Both metrics are obtained using  filtered approach i.e, for each test triplet, we generate all potential triplets but ensuring that none of the generated triplet appears in training, validation or test triplets, and rank the test triplets to obtain MRR and Hits@K.  

The results of link prediction on all datasets are shown in Table~\ref{tab:res}. Our model ~\relatt consistently shows performance improvements in all three datasets across all baselines. Our model (\relatt) shows an average improvement of 2.8\% in \textit{MRR} in all three datasets. We attribute this to the attention mechanism, that helps in learning weights for the messages from the neighbors. This suggests that all neighbors are not equal, and have different roles.

\begin{table}[!h]
\centering
\caption{Results of link prediction on FB15k-237, WN18 and \szat dataset.}
\label{tab:res}
\begin{tabular}{ccccc}
\toprule
 Dataset & Method                   & \distmult  & \rgcn   & \relatt  \\ 
 \midrule
\multirow{3}{*}{FB15k-237}&  MRR    & 0.201     &  0.226  & \textbf{0.234} \\ 
                        & Hits@1    & 0.130     & 0.126     &  \textbf{0.139} \\
                        & Hits@3    &  0.219    & 0.252     &  \textbf{0.258}   \\
                        & Hits@10   &  0.363    & 0.421     &  \textbf{0.428}  \\
 \midrule
 \multirow{3}{*}{WN18}  &  MRR      & 0.745     &  0.750    &  \textbf{0.767} \\ 
                        & Hits@1    & 0.592     &  0.612     & \textbf{0.639}  \\
                        & Hits@3    & \textbf{0.892}    &  0.882     & 0.889   \\
                        & Hits@10   &  0.934    &  0.939    &  \textbf{0.940}  \\
 \midrule
 \multirow{3}{*}{\szat}&  MRR       & 0.081     & 0.110     & \textbf{0.113}  \\ 
                        & Hits@1    & 0.046     & \textbf{0.073}     & \textbf{0.073}  \\
                        & Hits@3    &  0.082    & 0.107     & \textbf{0.113}    \\
                        & Hits@10   &  0.142    & 0.173     & \textbf{0.178}   \\
\bottomrule
\end{tabular}
\end{table}

\subsection{Unsupervised Task: Entity Matching}
\label{subs:entitym}
Next, we demonstrate the effectiveness of \relatt on learning a useful graph representation in an unsupervised setting. We evaluate \relatt with an entity matching task which aims to find data instances that refer to the same real-world entity. In particular, our defined task is to match entities across two knowledge graphs at a time. One graph is referred as a \textit{reference} graph which usually captures the full knowledge of a database. Given an entity in another graph referred as a query graph, the task is to find the corresponding entity in the reference graph.

We approach the above task in the following way: (1) we train \relatt on the full reference graph with the optimization goal to minimize the link prediction error; (2) we obtain the embeddings of the query entities by applying the trained \relatt on the query graphs; (3) we then perform matching, i.e. finding the most similar entities to the given entities in the query graphs by comparing the cosine similarity.


\begin{table}[]
\caption{Statistics of the query graphs}
\label{tab:qdata}
\begin{tabular}{cc}
\toprule
\textbf{Property} & \textbf{Measurement} \\
\midrule
\# of graphs       & 1,426 \\
avg nodes     & 20.90 \\
avg edges &  39.76 \\
\# of node types     &  4 \\
\# of relations     &   3 \\
\# of features      &  768 \\
\# of labels &  1,426     \\
avg node degree of the matching entity in R   &   19.88\\
avg node degree of the matching entity in Qs   &   12.89\\
\bottomrule
\end{tabular}
\end{table}

\subsubsection{Dataset}
The dataset consists of a reference graph and a set of query graphs. We use the full \szat as the reference graph to train \relatt. The query graphs are extracted from a proprietary relational database and the ground truth of the correct matching are manually labeled. The statistics of the dataset is reported in Table~\ref{tab:qdata}. From the query graph, we obtained multiple variations of query graphs by sampling neighbors of a matching entity based on their average node degree. We used a threshold ($th$) to filter out the neighbors. For example,  at $th=0.2$ we select 20\% of neighbors and  the average node degrees of query graphs reduce to 9.8.
These variations also limit the contextual information. A lower value of threshold produces query graphs with lesser contextual information. The idea is to test the model behaviour when query graphs have less contextual information present. 

\subsubsection{Baseline}
We evaluate our model against two methods. The first baseline uses only textual attributes present in the nodes of the query graphs. These textual features are fed into pre-trained BERT~\cite{bert} model to obtain the embeddings of entities. Second, baseline \rgcn ingests both attributes and query graph structure to obtain the embeddings. 
\subsubsection{Experimental Setup and Results}
We followed the same training protocol described in subsection~\ref{subs:expsetup} to train \relatt and \rgcn models. We use this trained model to infer the embeddings of entities in query graphs. Now having the embeddings of entities in knowledge graph and inferred embedding of query node,  we use cosine similarity metric to find a similar entity to the query node in the knowledge graph, and report the results on \textit{Hits@K}  in Table~\ref{tab:match}. The results show that using only textual attributes is not sufficient for entity matching as it does not use any contextual information. It can be seen in the case of \rgcn, as value of $th$  increases from 0.2 to 0.5, the performance of \rgcn improves because of the increased amount of context information. At lower values of contextual information (0.2), \rgcn  performance gets worse as compared to BERT, possibly because \rgcn does not differentiate between neighbors. This disadvantage is overcome by \relatt as it employs an attention mechanism that helps in learning better representations. \relatt performs better even when less contextual information is present. 
 
\begin{table}[]
\centering
\caption{Results of entity matching on \szat dataset.}
\label{tab:match}
\begin{tabular}{cccccc}
\toprule
     \textit{th}                                     & \textbf{Model}  & \textbf{Hits@1} & \textbf{Hits@5} & \textbf{Hits@10} & \textbf{Hits@30} \vspace{3pt}\\
\midrule
                 -                          & BERT           &  0.7373 &  0.8200 & 0.8440 & 0.8759\vspace{3pt} \\
\midrule
\multicolumn{1}{c}{\multirow{2}{*}{0.2}} & \rgcn          &  0.6248 & 0.7061  & 0.7446  & 0.7845\\ 
\multicolumn{1}{c}{}                     & \relatt        &  \textbf{0.7380} & \textbf{0.8425}  & \textbf{0.8766}  & \textbf{0.9231} \vspace{3pt} \\
\midrule
\multirow{2}{*}{0.3}                       & \rgcn          &  0.6270 & 0.7170  & 0.7685  & 0.8324\\ 
                                           & \relatt        &  \textbf{0.7540} & \textbf{0.8549}  & \textbf{0.8846} &\textbf{0.9253}\vspace{3pt}\\
\midrule
\multirow{2}{*}{0.4}                       & \rgcn           &  0.6567 & 0.7888  & 0.8440  & 0.8977\\ 
                                           & \relatt        &  \textbf{0.7627} & \textbf{0.8621}  & \textbf{0.8875} & \textbf{0.9209} \vspace{3pt}\\
\midrule
\multirow{2}{*}{0.5}                       & \rgcn          &  0.7250 & 0.8520  & 0.8861 & 0.9202\\ 
                                           & \relatt        &  \textbf{0.7736} & \textbf{0.8665} & \textbf{0.8919} & \textbf{0.9245}\\
\bottomrule
\end{tabular}
\end{table}





\section{conclusion}
\label{sec:conc}
In this work, we proposed a relation-aware masked attention mechanism that leverages the relation and neighborhood information to compute the importance of neighbors. Using this attention, the features are propagated from the neighbors of an entity to update its embedding. We evaluated the proposed model on the link prediction task on three datasets and entity matching task on one dataset which showed that the attention mechanism helps in learning a better representation.

This work  uses the KG structure and attributes present on the entities, and treats the entities as homogeneous types. This opens up the future direction to exploit the heterogeneity of entities in learning. Another exciting direction is to explore the sampling strategies on KGs such that the computational costs of training on large graphs can be reduced without losing the quality of learned embeddings. 

\bibliographystyle{ACM-Reference-Format}
\bibliography{bibliography}
\end{document}